# Implementación de Navegación en Plataforma Robótica Móvil Basada en ROS y Gazebo

Da Silva Angel, angel.dasilva@estudiantes.utec.edu.uy[1]
Fernández Santiago, santiago.fernandez@estudiantes.utec.edu.uy[1]
Vidal Braian, braian.vidal@estudiantes.utec.edu.uy[1]

Sodre Hiago, hiago.sodre@utec.edu.uy[1]
Moraes Pablo, pablo.moraes@utec.edu.uy[1]
Peters Christopher, qristopherp@gmail.com[2]
Barcelona Sebastian, sebastian.barcelona@utec.edu.uy[1]
Sandin Vincent, vincent.sandin@estudiantes.utec.edu.uy[1]
Moraes William, william.moraes@estudiantes.utec.edu.uy[1]
Mazondo Ahilen, ahilen.mazondo@estudiantes.utec.edu.uy[1]
Macedo Brandon, brandon.macedo@estudiantes.utec.edu.uy[1]
Assunção Nathalie, nathalie.assuncao@utec.edu.uy[1]
de Vargas Bruna, bruna.devargas@utec.edu.uy[1]
Kelbouscas André, andre.dasilva@utec.edu.uy[1]
Grando Ricardo, ricardo.bedin@utec.edu.uy[1]

[1]Universidad Tecnológica del Uruguay
[2]Ostfalia University of Applied Sciences

**Abstract:** *This research focused on utilizing ROS2 and Gazebo for simulating the TurtleBot3 robot, with the aim of exploring autonomous navigation capabilities. While the study did not achieve full autonomous navigation, it successfully established the connection between ROS2 and Gazebo and enabled manual simulation of the robot's movements. The primary objective was to understand how these tools can be integrated to support autonomous functions, providing valuable insights into the development process. The results of this work lay the groundwork for future research into autonomous robotics. The topic is particularly engaging for both teenagers and adults interested in discovering how robots function independently and the underlying technology involved. This research highlights the potential for further advancements in autonomous systems and serves as a stepping stone for more in-depth studies in the field.*

**Resumen:** *Esta investigación se centró en utilizar ROS2 y Gazebo para simular el robot TurtleBot3, con el objetivo de explorar las capacidades de navegación autónoma. Aunque el estudio no logró implementar la navegación autónoma completa, se estableció con éxito la conexión entre ROS2 y Gazebo y se permitió la simulación manual de los movimientos del robot. El objetivo principal era entender cómo estas herramientas pueden integrarse para soportar funciones autónomas, proporcionando conocimientos valiosos sobre el proceso de desarrollo. Los resultados de este trabajo sienta las bases para futuras investigaciones en robótica autónoma. El tema resulta particularmente interesante tanto para adolescentes como para adultos que desean descubrir cómo funcionan los robots de manera independiente y la tecnología subyacente involucrada. Esta investigación destaca el potencial para avances adicionales en sistemas autónomos y sirve como un punto de partida para estudios más profundos en el campo.*

Palabras clave: *ROS2, Gazebo, Simulación, TurtleBot, Robots Autónomos, Navegación Autónoma.*

## 1 - INTRODUCCIÓN

La capacidad de los robots para moverse y actuar de manera autónoma es fundamental en varias industrias. En logística, por ejemplo, estos robots mejoran significativamente la eficiencia operacional. Pereira et al. (2022), ilustra la relevancia de la inteligencia artificial en los sistemas industriales, demostrando cómo la automatización y la robótica móvil son clave para la innovación y eficiencia en este sector. Estos robots pueden transportar mercancías, recoger y entregar productos pesados, y gestionar inventarios de manera precisa. Esto no solo optimiza el flujo de trabajo, sino que también reduce errores y disminuye el riesgo de lesiones asociadas con el manejo manual de materiales. La navegación autónoma en estos robots permite una operación continua y sin interrupciones, siendo esencial para maximizar la productividad y seguridad en el sector.

Además de su impacto en la logística, los robots autónomos también revolucionan la conducción. Los vehículos equipados con sistemas de conducción autónoma, que integran tecnologías avanzadas para la navegación y la toma de decisiones, pueden ofrecer trayectorias más seguras y rápidas en comparación con la conducción humana. En este contexto, el Turtlebot 3 y ROS2 son herramientas clave para la investigación y desarrollo de navegación autónoma. ROS2 proporciona un marco robusto para implementar y simular algoritmos de navegación, mientras que el TurtleBot 3 permite probar estos sistemas en entornos controlados, ayudando a avanzar hacia un transporte más seguro y eficiente.

Otro campo en el que los robots autónomos tienen un impacto significativo es la exploración espacial. Al igual que en la logística y la conducción, la capacidad de los robots para operar de manera autónoma se revela crucial en este ámbito. Algunos trabajos ya exploran el uso de vehículos del tipo comentado, como el ejemplo donde se presenta un vehículo aéreo capaz de ser manejado autónomamente por visión computacional. Grando et al. (2020). Los robots espaciales pueden estudiar terrenos desconocidos y recolectar datos valiosos sin poner en riesgo la vida humana. Diseñados para realizar tareas complejas en ambientes extremos, estos robots previenen accidentes y complicaciones durante las misiones espaciales. A medida que la humanidad explora la posibilidad de vida en otros planetas y realiza descubrimientos en ellos, contar con robots especializados para estas tareas se vuelve esencial, demostrando así la versatilidad y la importancia de la autonomía robótica en diversas aplicaciones.

En el contexto actual de la robótica y la automatización, este trabajo se enfoca en enfrentar el desafío del manejo autónomo de un robot específico, el Turtlebot3. El objetivo principal de este estudio es lograr una simulación efectiva del Turtlebot3 operando de manera autónoma.

## 2 - REFERENCIAL TEÓRICO

Para comprender el desafío que presenta la simulación de un robot, es esencial tener una base sólida en las herramientas y tecnologías involucradas. Esto incluye el conocimiento sobre ROS, Gazebo, Linux, y las aplicaciones específicas de ROS. A continuación, se detallan estos elementos fundamentales. ROS2 (Robot Operating System 2) es la segunda generación del sistema operativo para robots desarrollado por la *Open Source Robotics Foundation* (OSRF).

ROS2 es una plataforma de código abierto diseñada para facilitar la creación y aplicación de sistemas robóticos. ROS2 ofrece múltiples ventajas, como la capacidad de funcionar en diferentes sistemas operativos, incluidos Linux, Windows y MacOS. Además, ROS2 incorpora más herramientas y paquetes que su predecesor, ROS1. También proporciona soporte para operaciones en tiempo real, una interfaz mejorada, y cuenta con una comunidad muy activa que contribuye continuamente a su desarrollo y expansión.

ROS2 es un sistema que se aplica en una amplia variedad de áreas que requieren robótica. Esto abarca desde la robótica móvil e industrial hasta vehículos autónomos, robótica de exploración y automatización del hogar, entre otros. Existen varias versiones de ROS2 que han sido lanzadas a lo largo del tiempo, cada una con mejoras y características específicas. Entre estas versiones se encuentran: ROS 2 Ardent Apalone (Diciembre, 2017), ROS 2 Bouncy Bolson (Mayo, 2018), ROS 2 Crystal Clemmys (Diciembre, 2018), ROS 2 Dashing Diademata (Mayo 2019), ROS 2 Eloquent Elusor (Noviembre 2019), ROS 2 Foxy Fitzroy (Junio 2020), ROS 2 Galactic Geochelone (Mayo, 2021), ROS 2 Humble Hawksbill (Mayo 2022), y ROS 2 Iron Irwini (Mayo 2023). La próxima versión, ROS 2 Jupiter Kicking Horse, está programada para ser lanzada en 2024.

Sin embargo, no todas estas versiones son compatibles con cada versión de Linux Ubuntu. En el caso de este estudio, la versión del software Ubuntu en uso es Ubuntu 22.04, y las versiones disponibles para esta plataforma son ROS 2 Iron Irwini y ROS 2 Humble Hawksbill. Otras versiones anteriores suelen estar disponibles para versiones anteriores de Ubuntu, como 18.04 o 20.04.

Gazebo es otro componente crucial para la simulación de robots. Es un simulador de robots y entornos tridimensionales también desarrollado por OSRF. Gazebo es ampliamente utilizado en la investigación y desarrollo de sistemas robóticos, así como en la industria para probar y validar algoritmos en un entorno virtual. Ofrece una variedad de funciones, como simulaciones realistas que incluyen físicas y entornos detallados, la capacidad de diseñar y personalizar entornos, y la posibilidad de visualizar datos de componentes y simular múltiples robots simultáneamente.

Por último, Linux es un sistema operativo de código abierto basado en Unix, creado por Linus Torvalds en 1991. Es conocido por su estabilidad, seguridad y alta personalización. Linux permite que varios usuarios trabajen en el mismo sistema simultáneamente y se distingue por su extensa capacidad de comandos, lo que proporciona una flexibilidad considerable para el manejo y la configuración del sistema operativo. La navegación autónoma de robots implica que estos sistemas sean capaces de moverse y realizar tareas sin intervención humana directa, basándose en su capacidad para percibir y entender su entorno. Para lograr una navegación autónoma efectiva, se utilizan diversos tipos de sensores que proporcionan información crucial sobre el entorno del robot, permitiendo tomar decisiones en tiempo real.

Entre los sensores más comunes se encuentran:
- **LIDAR (Light Detection and Ranging):** Utiliza pulsos de luz láser para medir distancias a objetos en el entorno. Calculando el tiempo que tarda el pulso en reflejarse y regresar al sensor, el LIDAR crea un mapa detallado en 3D del entorno, siendo fundamental para la creación de mapas precisos y la detección de obstáculos.

- **Cámaras:** Capturan imágenes del entorno en diferentes espectros (visible, infrarrojo, etc.). Estas imágenes son procesadas para extraer información relevante sobre los objetos, su posición, y el entorno general, facilitando el reconocimiento de objetos, la detección de señales, y la navegación visual.

- **Sensores de Colisión (Bumpers):** Se colocan en la parte exterior del robot y detectan el contacto físico con obstáculos. Normalmente mecánicos, estos sensores se activan al tocar un objeto, previniendo colisiones y ayudando al robot a detenerse o cambiar de dirección.

- **Sensores Ultrasónicos:** Emplean ondas sonoras de alta frecuencia y miden el tiempo que tardan en reflejarse de vuelta al sensor para calcular la distancia a objetos cercanos, siendo útiles para detectar obstáculos y medir distancias en entornos próximos.

- **Sensores Inerciales (IMU - Unidad de Medida Inercial):** Combinan acelerómetros y giroscopios para medir la aceleración y rotación del robot. Esto es esencial para estimar su orientación y movimiento, y es crucial para la estabilización y el control del movimiento del robot.

Cada tipo de sensor ofrece ventajas y limitaciones propias. Por ello, muchos sistemas de navegación autónoma combinan varios tipos de sensores para obtener una visión más completa del entorno. La integración de datos de múltiples sensores permite al robot tomar decisiones más informadas y adaptarse de manera más efectiva a las diversas condiciones del entorno, mejorando así su eficacia y seguridad en la navegación.

Es sumamente necesario comprender mínimamente los robots utilizados, como puede ser sus funciones e implementaciones. Los TurtleBots son plataformas robóticas móviles de código abierto, ampliamente utilizadas en el ámbito educativo, en la investigación, y en el desarrollo de aplicaciones robóticas.

El primer Turtlebot 3 se construyó utilizando un iRobot Create y un sensor Kinect de Xbox, lo que le permitió contar con una visión tridimensional del entorno (Marder-Epstein et al., 2010). Esta plataforma permitía a los usuarios mapear su entorno, como una casa, y programar al robot para que se desplazara de manera autónoma entre diferentes habitaciones. Además, los usuarios podían desarrollar aplicaciones para capturar imágenes de 360 grados o seguir a una persona en movimiento. Con el tiempo, los TurtleBots han evolucionado para ser más pequeños, asequibles y personalizables. En otras palabras, se optó por usar este robot en la investigación por el motivo de que es práctico y agradable en su uso. Además que sus componentes se adecuaban a la simulación propuesta.

Sus ventajas son Código Abierto y ROS, Modularidad y Personalización, Accesibilidad, Aplicaciones Versátiles.

Un Turtlebot3 incluye los siguientes componentes:

- 2x motores Dynamixel

- Sensor Lidar
- Raspberry Pi 3
- OpenCR
- Baterías Li-Po

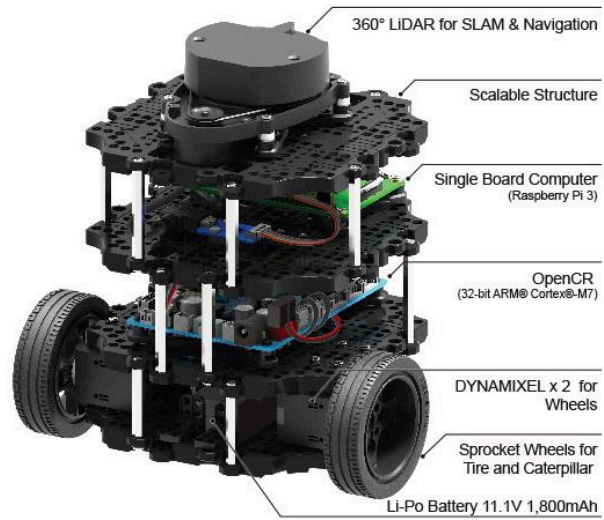

Figura 1. Partes de un Turtlebot3. ( Robotis. (n.d.). *Turtlebot3 Burger components*, 2024).

## 3 - METODOLOGÍA

La investigación tenía como meta crear un ambiente de simulación que pudiera recibir comandos de movimiento y generar un terreno adecuado para la práctica. A medida que avanzó el estudio, este trabajo también proporcionó capacitación en el uso del sensor LIDAR, permitiendo explorar su implementación en la simulación. Esto representa un desafío significativo para investigaciones futuras.

Arquitectura de Agentes en ROS 2:

- Nodos: Son procesos individuales que ejecutan tareas específicas en un sistema robótico. En una simulación con Turtlebot3, un nodo podría controlar el movimiento del robot, mientras que otro gestiona las lecturas del LiDAR. La modularidad de los nodos permite una arquitectura flexible y escalable.
- Tópicos: Son canales de comunicación que permiten a los nodos intercambiar mensajes. Por ejemplo, los datos del LiDAR se publican en un tópico, y otros nodos que necesitan esos datos para generar un mapa se suscriben a ese tópico.
- Servicios: Permiten la comunicación sincrónica entre nodos, útil para operaciones que requieren una respuesta inmediata, como comandos directos al Turtlebot3.
- Acciones: Ofrecen una forma de comunicación asíncrona con retroalimentación continua, ideal para tareas que requieren monitoreo en tiempo real, como la navegación autónoma del robot.
- Paquetes: Agrupan nodos, bibliotecas y archivos de configuración necesarios para la aplicación. En la simulación de un Turtlebot3, un paquete podría contener todo lo necesario para realizar el escaneo del entorno y generar un mapa.

Este enfoque modular y comunicativo de ROS 2 es esencial para la simulación y el escaneo de mapas en Gazebo con un TurtleBot3, utilizando su LiDAR para obtener datos precisos del entorno.

Los tópicos que se utilizaron para ejecutar la simulación del Turtlebot3 fueron los siguientes: cmd_vel, odom, gazebo, entre otros como los siguientes:

```
$ ros2 launch turtlebot3_gazebo turtlebot3_dqn_stage4.launch
$ ros2 launch turtlebot3_cartographer cartographer.launch.py
```

```
$    ros2    launch    turtlebot3_navigation2    navigation2.launch.py
use_sim_time:=True map:=$HOME/mapp.yaml
$ ros2 run nav2_map_server map_saver_cli -f ~/mapp
$ ros2 run turtlebot3_teleop teleop_keyboard
```

El primero de ellos implementa el modelo del robot a utilizar en Gazebo, la siguiente línea de comandos permite activar las funcionalidades del LIDAR y con esto la detección de obstáculos, la tercer línea abre el mapa guardado pero para ello tiene que estar abierto Gazebo e inicializado el comando Turtlebot3 cartographer. La cuarta línea de comandos, permite salvar/guardar el mapa creado en la carpeta que se desee. Y el último de ellos, permite la conexión entre el robot simulado y el teclado del computador, con el fin de ser manejado manualmente.

Aunque claramente, la lista de comandos a ofrecer era mucho más amplia, algunos de ellos son:

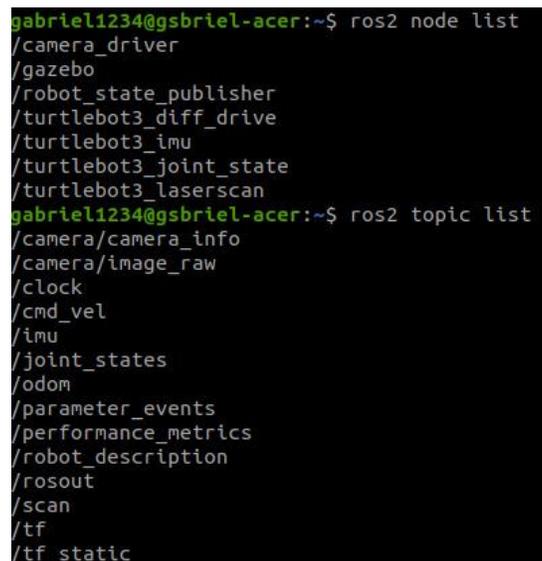

Figura 2. Lista de nodos y tópicos disponibles.

## 4 - RESULTADOS

Los resultados obtenidos cumplieron con los objetivos propuestos, donde el robot Turtlebot3 simulado logró la conexión y las acciones solicitadas, permitiendo su movimiento manual mediante el teclado del computador. Además, se logró modificar el mapa en Gazebo y establecer su conexión con ROS 2. La estrategia utilizada fue diseñar un mapa adecuado con los obstáculos necesarios para que el robot no pudiera completar el trayecto de manera sencilla.

Posteriormente, se identificaron los comandos necesarios para conectar con Gazebo y crear dicho mapa. Se llevaron a cabo diversas pruebas en cada uno de los mapas para encontrar el más óptimo. Las Figuras 3, 4 y 5 presentan estos resultados.

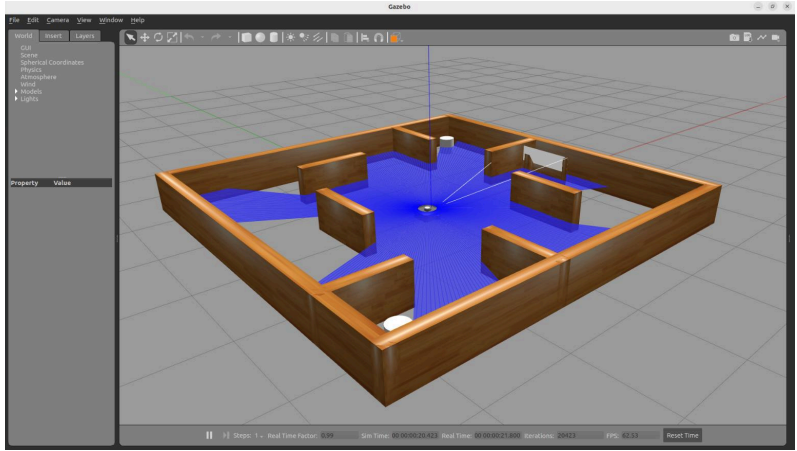
Figura 3. Primer Mapa Creado.

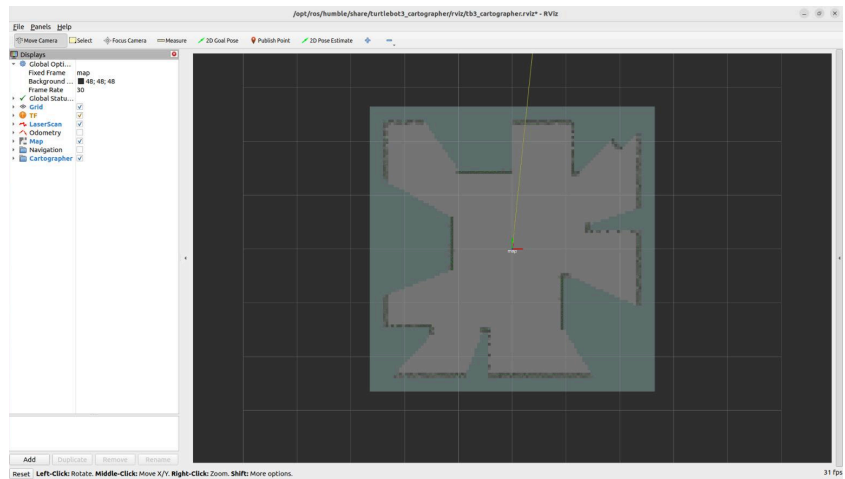
Figura 4. Segundo Mapa Creado.

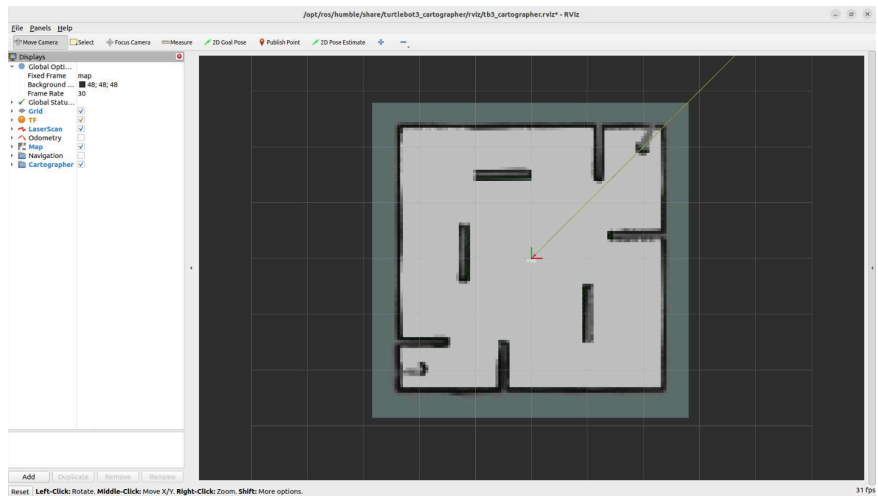
Figura 5. Mapa Final.

Se testeó y alcanzó el movimiento del robot simulado manualmente, modificando los valores de velocidad del robot. A futuro se pueden implementar algoritmos de movimiento y ejecuciones de comandos al robot vía programación, lo que, con esta plataforma de simulación, torna posible la mejor integración. A continuación es posible visualizar en las Figuras 4 y 5 algunos resultados de la simulación en Gazebo y Rviz.

Además de los anteriores avances, fue posible la práctica con el sensor LIDAR, logrando la detección completa de diferentes obstáculos del mapa. También se guardó el mapa escaneado en un archivo para a futuro desarrollar el funcionamiento del robot de forma autónoma, obteniendo conocimiento valioso para una futura investigación.

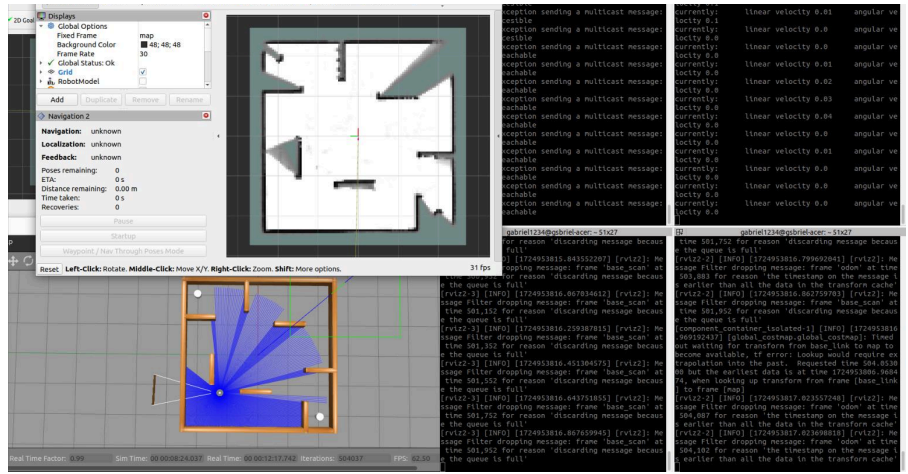

Figura 6. Ambiente de simulación Gazebo para el Turtlebot3.

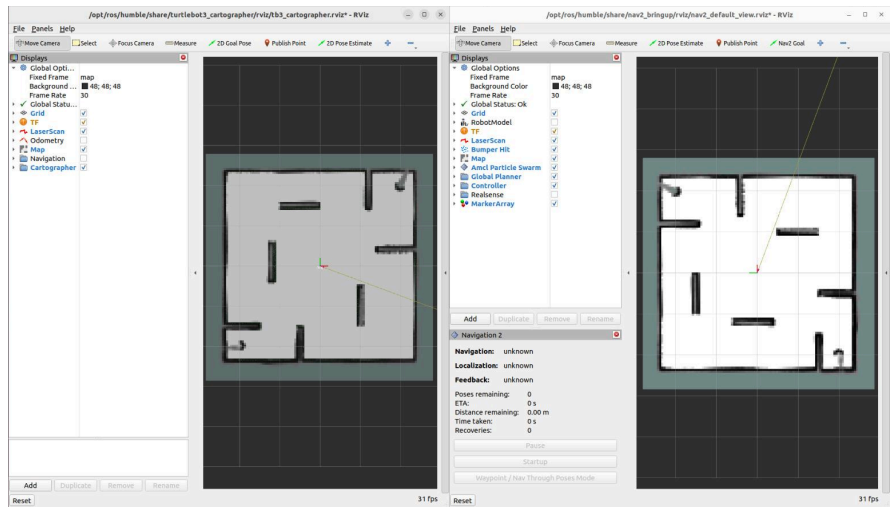

Figura 7. Funcionamiento del LIDAR en la Simulación en Gazebo.

Como se ve en la figura 7, se consiguió comprender el funcionamiento del LIDAR y su implementación, lo que nos permite generar nuevos entrenamientos con la finalidad de obtener la navegación autónoma.

**5 - CONCLUSIONES**

El trabajo se propuso alcanzar la conexión entre ROS2 y el TurtleBot3, generar un mapa con diversos obstáculos en Gazebo, reconocer estos obstáculos y crear un mapa cargado con ellos. Instalación de bibliotecas necesarias para la simulación, el movimiento completamente manual del robot y el aprendizaje para realizar su movimiento de forma autónoma en un futuro.

Los resultados obtenidos cumplieron con las expectativas: se alcanzaron todos los objetivos. Establecimos la conexión entre el robot y ROS 2, creamos mapas con obstáculos que dificultan el movimiento del robot para luego realizar la simulación. También guardamos el mapa después de ser escaneado e instalamos todas las bibliotecas necesarias. Se logró el movimiento manual del robot TurtleBot3 utilizando el teclado del computador, lo que permitió modificar los datos de velocidad. Además, se adquirió nuevo conocimiento sobre el uso de IA para el movimiento autónomo del robot, enfocándose en la evasión de obstáculos y la selección del mejor camino mediante sensores como el LIDAR.

A futuro se busca implementar dicha capacidad de moverse autónomamente mediante el uso de sensores como LIDAR, al lograrlo, se buscará conectar de forma remota con el Turtlebot3 físico y de esta manera entrenarlo con el fin de que esquive obstáculos de determinada área.

Además claro, de intentar el uso de nuevos sensores en el robot y de esta manera generar un estudio del terreno mucho más claro.

**6 - REFERENCIAS**